\def\paperID{970}
\def\confName{CVPR}
\def\confYear{2024}

\newif\ifreview 
\newif\ifarxiv \newcommand{\arxiv}{\arxivtrue}
\newif\ifcamera 
\newif\ifrebuttal

\arxiv
\documentclass[10pt,twocolumn,letterpaper]{article}
\ifreview \usepackage[review]{cvpr} \fi
\ifarxiv \usepackage[pagenumbers]{cvpr} \fi
\ifrebuttal \usepackage[rebuttal]{cvpr} \fi
\ifcamera \usepackage{cvpr} \fi

\usepackage{xr-hyper}

\makeatletter
\newcommand*{\addFileDependency}[1]{
  \typeout{(#1)}
  \@addtofilelist{#1}
  \IfFileExists{#1}{}{\typeout{No file #1.}}
}

\makeatother

\usepackage[pagebackref,breaklinks,colorlinks,citecolor=cvprblue]{hyperref}
\usepackage[capitalize]{cleveref}
\crefname{section}{Sec.}{Secs.}
\crefname{table}{Table}{Tables}
\crefname{figure}{Fig.}{Figs.}

\usepackage{booktabs}       %
\usepackage{xcolor}         %
\usepackage{float}
\usepackage{comment}

\usepackage{multirow}
\usepackage{tabularx}
\usepackage{mathtools}
\usepackage{booktabs}

\usepackage{enumitem}%
\usepackage{verbatim}
\DeclareMathSymbol{\shortminus}{\mathbin}{AMSa}{"39}
\DeclareFontFamily{OT1}{pzc}{}
\DeclareFontShape{OT1}{pzc}{m}{it}{<-> s * [1.150] pzcmi8t}{}
\DeclareMathAlphabet{\mathpzc}{OT1}{pzc}{m}{it}
\usepackage{soul}
\usepackage{xfakebold}

\newcommand{\fbseries}{\unskip\setBold\aftergroup\unsetBold\aftergroup\ignorespaces}
\makeatletter
\newcommand{\setBoldness}[1]{\def\fake@bold{#1}}
\makeatother

\definecolor{cadmiumgreen}{rgb}{0.0, 0.42, 0.24}

\DeclareMathSymbol{\shortminus}{\mathbin}{AMSa}{"39}

\newcommand{\fig}[1]{Fig~\ref{fig:#1}}
\newcommand{\sect}[1]{Sect~\ref{sect:#1}}

\newcommand{\tab}[1]{Table~\ref{tab:#1}}

\newcommand{\eq}[1]{Eq. (\ref{eq:#1})}

\usepackage{xspace}
\newcommand{\MethodName}{Rig3DGS\xspace}

\newcommand{\xbf}{\mathbf{x}}
\newcommand{\vbf}{\mathbf{v}}

\newcommand{\etabf}{\pmb{\eta}}
\newcommand{\gammabfiexp}{\pmb{\gamma}_{i,\text{exp}}}
\newcommand{\gammabfipose}{\pmb{\gamma}_{i,\text{pose}}}

\usepackage{environ}
\NewEnviron{smequation}{%
    \begin{equation}
    \scalebox{0.85}{$\BODY$}
    \end{equation}
}
\title{Rig3DGS: Creating Controllable Portraits from Casual Monocular Videos}

\author{%
Alfredo Rivero* \\
  Stony Brook University \\
  \texttt{alrivero@cs.stonybrook.edu}
  \and
  ShahRukh Athar* \\
  Stony Brook University \\
  \texttt{sathar@cs.stonybrook.edu}
  \and
  Zhixin Shu \\
  Adobe Research \\
  \texttt{zshu@adobe.com} \\
   \and
   Dimitris Samaras \\
  Stony Brook University \\
  \texttt{samaras@cs.stonybrook.edu}
}

\definecolor{cvprblue}{rgb}{0.21,0.49,0.74}
\begin{document}

\maketitle
\def\thefootnote{*}\footnotetext{Equal Contribution}\def\thefootnote{\arabic{footnote}}

\begin{abstract}

Creating controllable 3D human portraits from casual smartphone videos is highly desirable due to their immense value in AR/VR applications. The recent development of 3D Gaussian Splatting (3DGS) has shown improvements in rendering quality and training efficiency. However, it still remains a challenge  to accurately model and disentangle head movements and facial expressions from a single-view capture to achieve high-quality renderings. In this paper, we introduce \MethodName to address this challenge. We represent the entire scene, including the dynamic subject, using a set of 3D Gaussians in a canonical space. Using a set of control signals, such as head pose and  expressions, we transform them to the 3D space with learned deformations to generate the desired rendering. Our key innovation is a carefully designed deformation method which is guided by a learnable prior derived from a 3D morphable model. This approach is highly efficient in training and effective in controlling facial expressions, head positions, and view synthesis across various captures. We demonstrate the effectiveness of our learned deformation through extensive quantitative and qualitative experiments. \href{http://shahrukhathar.github.io/2024/02/05/Rig3DGS.html}{The project page can be found here.}

\end{abstract}

\section{Introduction}
Creating controllable 3D human portraits is crucial for various immersive experiences, including virtual reality, telepresence, film production, and educational applications. Yet, the realization of this technology by everyday consumers using only basic smartphone cameras presents considerable challenges.
Modeling a 3D controllable portrait from videos typically involves either an explicit or implicit registration of the dynamic human subject, accounting for varying factors, such as facial expressions and head poses in each frame. This process requires a precise disentanglement of the facial deformations caused by these factors which is often challenging without ground truth. The challenge is further compounded when using a monocular capture, as each head pose and expression is only visible from a single viewpoint, making accurate disentanglement more complex.

While there has been some prior work, such as RigNeRF \cite{athar2022rignerf}, to address these challenges, it is slow to both train and render. This is due to the use of a multi-layer perceptron (MLP) based neural radiance field (NeRF) to model the 3D scene. The poor representation capacity of these MLPs, even with positional encoding, leads to blurry results, especially for novel expressions and poses. More recent 3D representations, such as 3D Gaussians \cite{kerbl3Dgaussians}, significantly improve the rendering quality while being orders of magnitude faster to both train and render than MLP-based NeRFs. However, current works on 3D Gaussian Splatting (3DGS) cannot represent dynamic content and are unsuitable for reanimation tasks. An initial effort towards dynamic 3D gaussians \cite{luiten2023dynamic}, uses a multi-camera setup to reconstruct a point-cloud of the dynamic scene at every time-step and establishes correspondence across the points for training. However, their method only reconstructs the dynamic scene, it cannot be reanimated.

In a monocular setting, reconstructing the scene's point cloud in the dynamic regions proves to be difficult \cite{schoenberger2016sfm} due to only a single view being available at every time-step. Consequently, \cite{luiten2023dynamic}'s approach cannot establish weak correspondences across time in the dynamic regions of the scene to reconstruct the scene. In this paper, we present \MethodName, a method that ``rigs'' 3D gaussians and enables the creation of reanimatable neural-gaussian portraits. Our method provides full control over facial expressions, head-pose, and enables novel-view synthesis of a modeled subject and their entire scene. Our method uses a point cloud of 3D gaussians in a canonical space that is initialized from COLMAP \cite{schoenberger2016sfm} for the static region and the FLAME mesh \cite{FLAME:SiggraphAsia2017} for the dynamic regions (the human subject) of the scene. These canonical 3D gaussians are then transformed to a facial expression and head-pose dependent deformed space before they are rendered via differentiable Gaussian Splatting. Our key insight is to restrict our network's deformation from canonical to the deformed space using a \textit{learnable} prior derived from a 3D face mesh. More specifically, we predict the deformation from a 3D canonical space to a deformed space for each 3D Gaussian as a weighted sum of the deformations of its closest vertices on a morphable model mesh. These learnable weights are optimized via photometric loss w.r.t to a ground truth image. By restricting each Gaussian's deformation to lie in a linear subspace of the vertex deformations, we can effectively regularize an otherwise ill-posed problem of learning a per-point deformation without any ground truth available. In \sect{aba_def_prior}, we show that this learnable prior is essential to better generalization to novel facial expressions and head-poses. Once trained, portraits created by \MethodName can be reanimated with full control over facial expressions and head-pose of the subject and novel view synthesis of the entire scene. 
In summary, our contributions are as follows:
\begin{itemize}
    \item We propose a novel deformation model learnt in the subspace of the deformation defined by a 3D morphable model. This enables us to generalize to novel facial expressions and head-poses during reanimation.
    \item We propose \MethodName, a method that rigs 3D gaussians to enable to creation of reanimatable portraits with full control over facial expressions, head-pose and novel view synthesis of the entire scene from a casually-captured \textit{monocular} smartphone video.
    
    \item We demonstrate significant improvement in rendering quality over prior work on neural portraits with novel facial expressions, head-poses and novel views synthesis of the entire scene while being 50 times faster than prior work due to representing the scene as 3D gaussians. 
\end{itemize}

\begin{figure*}
    \centering
    \includegraphics[width=1.0\linewidth]{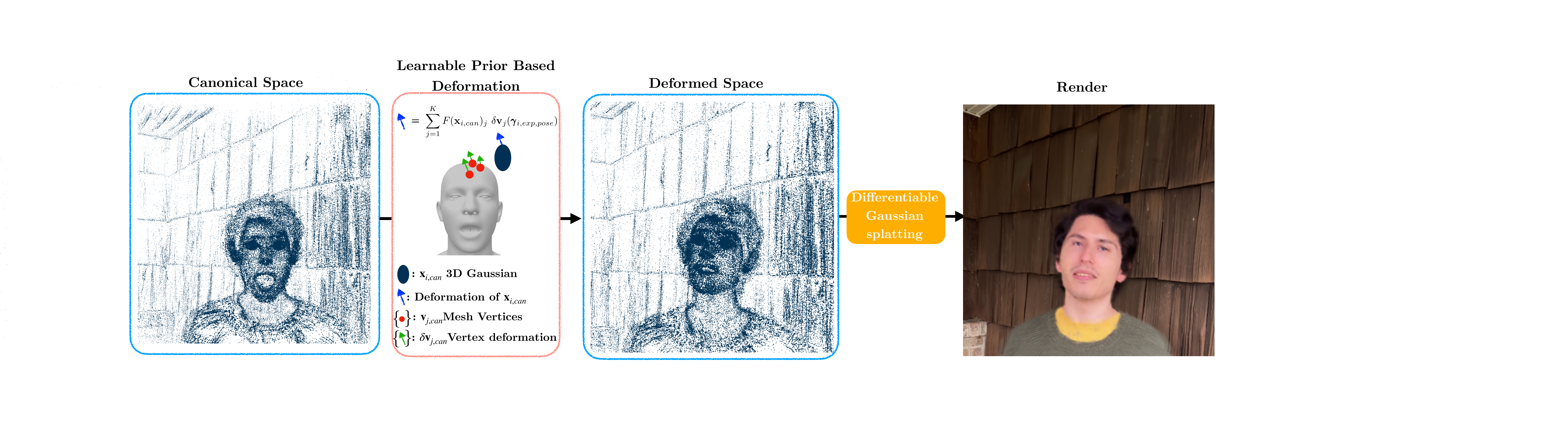}
    \caption{{\textbf{\MethodName.} Our method models the dynamic scene as a collection of 3D Gaussians in the canonical space that are deformed according to the target facial expression and head-pose to the deformed space before being rendered via differentiable splatting. We constrain the deformation to lie in the sub-space of local vertex deformation, which allows us to generate photorealistic renders with high fidelity to the target expression and head-pose. 
    }}
    \label{fig:method}
\end{figure*}

\section{Related Work}
\MethodName is a method for arbitrary facial expression control and novel view synthesis of scenes captured in portrait videos. It is closely related to recent work on neural rendering and novel view synthesis, 3D facial modeling, and controllable face generation.

\paragraph{Neural Scene Representations and Novel View Synthesis.} \MethodName is related to recent advances in neural rendering and novel view synthesis  \citep{nerf, nerfies, barron2022mipnerf360, yu2022plenoxels, mueller2022instant, Chen2022ECCV, kerbl3Dgaussians, DNeRF, liu2022devrf, jang2022dtensorf, luiten2023dynamic, INSTA:CVPR2023, Zheng2023pointavatar}. Neural Radiance Fields (NeRFs) learn a volumetric representation of a scene which typically, when provided a 3D point and the direction from which the point is being viewed, predicts color and volume density using a differentiable renderer. Our method is built upon 3D Gaussian Splatting \citep{kerbl3Dgaussians} and uses a dense point cloud of three-dimensional gaussian kernels to represent a scene's geometry. For any given camera pose, 3D gaussian kernels are projected onto the image evaluated densely pixel-by-pixel. Gaussian kernels within the local neighborhood of each pixel are sorted and pixel color is approximated using a modified approach to conventional $\alpha$-blending. 3D Gaussian Splatting and other NeRFs are trained by minimizing the error between the predicted color of a pixel and its ground truth value. While NeRFs are able to generate high quality and photo-realistic images for novel view synthesis, many are designed for static scenes and are unable to represent scene dynamism.

\paragraph{Dynamic Neural Scene Representations.} Methods such as \citep{DNeRF, liu2022devrf, D-TensoRF, luiten2023dynamic} extend NeRFs to dynamic scenes through specialized frame-to-frame parameter residual learning \citep{luiten2023dynamic} or incorporating a time component and canonical deformation network \citep{DNeRF, liu2022devrf, jang2022dtensorf}. Among approaches incorporating a canonical deformation network, a dynamic scene is decomposed into canonical and deformed spaces, with each moment in time defining a reconfiguration of the canonical space using learned displacements in position, color, and other parameters. Recent approaches to human body and facial reconstruction \citep{INSTA:CVPR2023, Zheng2023pointavatar, athar2022rignerf} refine and expand existing deformation-based NeRFs for their data domains, aiming to reconstruct and reanimate data with the assistance of existing templates and/or statistical models. %

\paragraph{Controllable Face Generation.} Recent breakthroughs in Generative Adversarial Networks(GANs) \citep{goodfellow2014generative, pix2pix2016, CycleGAN2017, StyleGAN, Karras-2019-ASB, Karras-2020-AAI} have enabled high-quality 2D image generation and manipulation. These 2D breakthroughs have inspired a large collection of work \citep{shu2018deforming, NeuralFace2017, athar2020self, pumarola2020ganimation, StarGAN2018, starganv2, tewari2020stylerig, tewari2020pie, deng2020disentangled, kowalski2020config} focusing on face image manipulation and editing. However, the majority of these works are intrinsically image-based and lack an explicit 3D representation. Therefore, these methods find it challenging to enable high-quality view synthesis and control portraits with large head pose changes or extreme facial expressions. Recent methods such as \citep{INSTA:CVPR2023, Zheng2023pointavatar, athar2022rignerf} address these shortcomings. Existing dynamic NeRF models that use 3D Morphable Models such as FLAME \cite{FLAME:SiggraphAsia2017}, INSTA \citep{INSTA:CVPR2023} and PointAvatar \citep{Zheng2023pointavatar} reconstruct and reanimate human heads and/or torsos at a high quality and with reasonable rendering/training times. However, these methods fail to capture an entire portrait scene, with both of these methods failing to capture a scene's background. Geometrically-complete approaches such as RigNeRF \citep{athar2022rignerf} exist, which incorporates a deformation prior defined by an existing FLAME mesh and corrected through residual learning. But, this method fails to reproduce facial portraits with reasonable rendering/training times.

\section{\MethodName}

In this section, we describe our method \MethodName, which enables novel view synthesis of neural portraits with arbitrary control over head pose and facial expressions. We represent the scene using 3D gaussians in the canonical space that are transformed to the deformed space using a learnable deformation prior. This prior restricts the deformation of a gaussian to lie in the subspace spanned by the deformation of its K-nearest vertices on the FLAME mesh \cite{FLAME:SiggraphAsia2017}. Similarly, the rotation of each gaussian is learnt as a correction to the rotation of its K-nearest vertices. An overview of our method is given in \fig{method}.    
\subsection{Preliminaries}
\subsubsection{3D Gaussian Splatting}
\label{sect:3dgs}
3D Gaussian Splatting works by representing a of three-dimensional gaussian `blob' with a learnable spatial extent, orientation, opacity and color. More specifically, each gaussian is defined by a covariance matrix  $\Sigma$ and mean \textbf{x} as \[G(x) = e^{-\frac{1}{2}(\textbf{x})^T \Sigma^{-1} (\textbf{x})}\] where $\Sigma$ is decomposed into two differentiable scaling and rotation matrices \textbf{S}, \textbf{R} such that \[\Sigma = \textbf{R}\textbf{S}\textbf{S}^T\textbf{R}^T.\] When rendering novel views, the gaussians are projected onto an image plane using a differentiable variant of Surface Splatting \cite{surfacesplatting}. Using the viewing transform matrix \(W\) and the Jacobian matrix \(J\) of the affine approximation of the projective transformation characterized by a novel view, a 3D gaussians's color and analogous 2D gaussian, whose covariance matrix $\Sigma'$ is defined as \[\Sigma'=JW \Sigma W^TJ^T,\] is blended with $N - 1$ neighboring 2D kernels to produce an arbitrary pixel's color. These kernels are blended using a modified approach to conventional $\alpha$-blending. The gaussians are sorted relative to their 3D mean's distance to the image plane, and a pixel's color $C$ is computed as \[C = \sum^{N}_{i=1}c_i \alpha_i \prod^{i - 1}_{j = 1} (1 - \alpha_j)\] where $c_i$ is the color of each point and $\alpha_i$ is given by evaluating a 2D gaussian with covariance $\Sigma'$ multiplied with each kernel's opacity.

\subsection{Deforming Gaussians with a Learnable Prior}
3D Gaussian Splatting, as described in \sect{3dgs}, does not support dynamic scene rendering, as the initial point-cloud is assumed to be static. In contrast, our scene is inherently dynamic with the subject constantly assuming different facial expressions and head-poses. We choose to model this dynamic behaviour as a deformation from a predefined canonical space where the 3D gaussians lie with positions \(\{\xbf_{i,can}\}\) to a deformed space with positions \(\{\xbf_{i,def}\}\) where they are rendered. The choice of the canonical frame is arbitrary, and we choose it to be defined by a close to neutral head-pose with an open mouth. The initial point cloud is a concatenation of the points on the background given by COLMAP \cite{schoenberger2016sfm} \(\{\xbf_{i,can}^{COLMAP}\}\) and the vertices of the FLAME mesh \cite{FLAME:SiggraphAsia2017} in the aforementioned canonical configuration  \(\{\vbf_{i,can}\}\). For brevity we will refer to the concatenated set of points as \(\{\xbf_{i,can}\} = \{\xbf_{i,can}^{COLMAP}\} \cup \{\vbf_{i,can}\}\). 
For any given frame \(i\) with expression and head-pose parameters \(\{\gammabfiexp,\gammabfiexp\}\), the deformation of each flame vertex is given as follows 
\begin{smequation}
\delta \vbf(\gammabfiexp,\gammabfiexp)_{j} = \vbf(\gammabfiexp,\gammabfiexp) - \vbf_{can}    
\label{eq:flame_vert}
\end{smequation}
where, \(\vbf(\gammabfiexp,\gammabfiexp)\) is the deformed position of the vertex for expression and head-pose parameters \(\{\gammabfiexp,\gammabfiexp\}\) calculated via FLAME \cite{FLAME:SiggraphAsia2017}. Next, for each gaussian in the canonical space,  \(\{\xbf_{i,can}\}\), we calculate its deformation using a linear combination of the deformation of it's nearest \(K\) vertices as follows  
\begin{smequation}
\begin{split}
    &\text{Def}(\xbf_{i,can}, \gammabfiexp,\gammabfipose) = \sum_{j=1}^{K}w_{j}\times \ \delta \vbf_{j}(\gammabfiexp,\gammabfipose) \\
    &\text{where} \quad \vbf_{j,can}\in kNN(\xbf_{i,can},\text{Mesh}, K) \text{ and }  \\
    & [w_{1},...,w_{K}] = \textbf{w} = \text{F}(\xbf_{i,can})  
\end{split}
\label{eq:deform_def_eq}
\end{smequation}
where, \(\text{Def}(\xbf_{i,can}, \gammabfiexp,\gammabfipose)\) is the deformation of the canonical gaussian at \(\xbf_{i,can}\), \(\{\vbf_{j,can}\} := kNN(\xbf_{i,can},\text{Mesh}, K) \) are the set of \(K\) closest mesh vertices to the gaussian at \(\xbf_{i,can}\) in the canonical space, \(w_{j}\) is the weight of vertex deformation \(\delta \vbf_{j}(\gammabfiexp,\gammabfipose)\) and \(F\) is a function that calculates these weights as a function of the canonical gaussian position \(\xbf_{i,can}\). We represent \(F\) using a triplane followed by a small 2-layer MLP. The gaussian at \(\xbf_{i,can}\) is then transformed to the deformed space as follows:

\begin{smequation}
\begin{split}
    \xbf_{i,def} = \xbf_{i,can} + \text{Def}(\xbf_{i,can}, \gammabfiexp,\gammabfipose)
\end{split}
\label{eq:def_eq}
\end{smequation}
 \eq{deform_def_eq} is flexible enough to model points on the static background by setting \(\textbf{w} = 0 \) while simultaneously being constrained enough on foreground points to lie in the subspace of vertex deformations when \(\textbf{w} \neq 0 \). In practice, subjects move a little during capture and the ground-truth deformations of the head can lie outside the subspace defined by the vertex deformations. To model these small deviations, we introduce global translation flexibility in \eq{deform_def_eq} 

\begin{smequation}
\begin{split}
    \text{Def}(\xbf_{i,can}, \gammabfiexp,\gammabfipose) &= \sum_{j=1}^{K}w_{j}\times \ \delta \vbf_{j}(\gammabfiexp,\gammabfipose) \\
    & + \eta + T(i)\\
     \text{where, } \quad \eta = \text{G}(\xbf_{i,can}, &\text{DWAvg}(\{\delta \vbf_{j}(\gammabfiexp,\gammabfipose\}\})))
\end{split}
\label{eq:deform_def_eq_final}
\end{smequation}

where,  \(\eta = \text{G}(.)\) is the corrective translation predicted by a small 2-layer MLP \(\text{G}\) that is conditioned the gaussian's canonical position \(\xbf_{i,can}\) and an inverse distance-weighted average of the deformation of vertices closest to it \( \text{DWAvg}(\{\delta\vbf_{j}(\gammabfiexp,\gammabfipose)\})\) and \(T(i)\) is a small per-frame deformation predicted using another small 2-layer MLP \(T\). In practice, we use \(K = 10\) nearest neighbors.

In order to ensure a smooth optimization and prevent local minima of the deformation we use some regularizations. First, we regularize  \(\text{Def}(\xbf_{i,can}, \gammabfiexp,\gammabfipose)\) to match the deformation given by FLAME at the mesh vertices as follows:
\begin{smequation}
    \mathcal{L}_{FLAME} =  \sum_{i=1}^{V}\||\text{Def}(\vbf_{i,can}, \gammabfiexp,\gammabfipose) - \delta \vbf_{i}(\gammabfiexp,\gammabfipose)\||_{2}^{2}
\end{smequation}
 where, \(V\) is the total number of mesh vertices. We also regularize the deformation to be close to zero for points that are further than a pre-defined threshold,\(D\) ,  of the mesh as measured in the canonical space:
\begin{smequation}
\begin{split}
    \mathcal{L}_{Global,Def} &= \||\text{Def}(\xbf_{l,can}, \gammabfiexp,\gammabfipose)\||_{2}^{2}; \\
    \text{where } \hspace{0.5em} &\{\xbf_{l,can} \text{ s.t } \text{ Dist}(\xbf_{l,can}, Mesh) >= D\}
\end{split}
\label{eq:def_reg}
\end{smequation}
Finally, we regularize the correction translation and per-frame deformation defined in \eq{deform_def_eq_final} to be small
\begin{smequation}
\begin{split}
    \mathcal{L}_{\eta} &= \||\etabf\||_{2}^{2}\\
    \mathcal{L}_{T} &= \||T(i)\||_{2}^{2}
\end{split}
\end{smequation} 
\subsection{Rotating and Scaling Gaussians}
As described in \sect{3dgs}, the gaussians we use to represent the scene have associated with them an orientation \(\textbf{R}\) and scaling \(\textbf{S}\) in addition to a canonical position. Thus, in addition to the translation defined in \eq{deform_def_eq_final}, we must also predict their rotation and scaling transformation to the deformed space. Similar to \eq{deform_def_eq_final}, we utilize the mesh to create a prior on the local rotation and predict a correction to it. More specifically, for any gaussian in the canonical space at position \(\xbf_{can}\) with rotation \(R_{can}\) in the canonical space, we predict the deformed space rotation \(R_{def}\) as
\begin{smequation}
\begin{split}
R_{def} = R_{can}&R'(\xbf_{i,can}, \delta \vbf_{j}(\gammabfiexp,\gammabfipose)); \\
    R'(\xbf_{i,can}, \delta \vbf_{j}&(\gammabfiexp,\gammabfipose)) \\
    & = R_{mesh}R^{*}(\xbf_{i,can}, \text{DWAvg}(\{\delta \vbf_{j}(\gammabfiexp,\gammabfipose)\}))\\
    \text{where } \hspace{0.5em}   \vbf_{j,can}&\in kNN(\xbf_{i,can},\text{Mesh}, K)
\end{split}
\end{smequation}
where, \(\{\vbf_{j,can}\} := kNN(\xbf_{i,can},\text{Mesh}, K) \) are the set of \(K\) closest mesh vertices to the gaussian at \(\xbf_{i,can}\) in the canonical space, \(R_{Mesh} = \texttt{Kabsch}(\vbf_{j,can}, \vbf_{j,def}(\gammabfiexp,\gammabfipose))\) is the rotation matrix between the aforementioned closest vertices in the deformed space and the canonical space calculated using the Kabsch algorithm \cite{kabsch} and \(R^{*}(\xbf_{i,can}, \{\delta \vbf_{j}(\gammabfiexp,\gammabfipose)\})\) is a rotation matrix predicted by a small MLP conditioned on the gaussian's canonical position and the inverse-distance weighted average of the deformations of the closest vertices. Similarly, we predict a deformed space scaling using an MLP, \(S^{*}(\xbf_{i,can}, \text{DWAvg}(\{\delta \vbf_{j}(\gammabfiexp,\gammabfipose))\})\), conditioned on the gaussian's canonical position and the inverse-distance weighted average of the deformations of the closest vertices as follows
\begin{smequation}
    \textbf{S}_{i,def} = \textbf{S}_{i,can}\times S^{*}(\xbf_{i,can}, \text{DWAvg}(\{\delta \vbf_{j}(\gammabfiexp,\gammabfipose))
\end{smequation}
Like in \eq{def_reg}, we regularize the rotation matrix to be identity for all points that are further than a pre-defined threshold,\(D\) ,  of the mesh as measured in the canonical space:
\begin{smequation}
\begin{split}
    \mathcal{L}_{Global,Rot} &= \||R(\xbf_{l,can}, \delta \vbf_{j}(\gammabfiexp,\gammabfipose)) - \mathbf{I}\||_{2}^{2}; \\
    \text{where } \hspace{0.5em} &\{\xbf_{l,can} \text{ s.t } \text{Dist}(\xbf_{l,can}, Mesh) >= D\}
\end{split}
\label{eq:def_reg_R}
\end{smequation}
We also regularize predicted scaling factor to be close to 1 for all points in the scene:
\begin{smequation}
    \mathcal{L}_{Global,scale} = \||S^{*}(\xbf_{i,can}, \text{DWAvg}(\{\delta \vbf_{j}(\gammabfiexp,\gammabfipose)) - 1\||
\end{smequation}
\subsection{Full Loss}
We supervise \MethodName using a photometric loss with respect to the ground-truth along with all other aforementioned regularizations on the deformation, the full loss is

\begin{smequation}
\begin{split}
    \mathcal{L} &= 0.8\times \mathcal{L}_{1} +  0.2\times \mathcal{L}_{D-SSIM} + \mathcal{L}_{FLAME} \\
    & + \lambda_{Global,Def}\mathcal{L}_{Global,Def} + \lambda_{\eta}\mathcal{L}_{\eta} + \lambda_{T}\mathcal{L}_{T} \\
    &+ \lambda_{Global,Rot}\mathcal{L}_{Global,Rot} + \lambda_{Global,scale}\mathcal{L}_{Global,scale} 
\end{split}
\end{smequation}
Empirically, we find the ideal weights are \( \lambda_{Global,Def}=1e-1, \lambda_{\eta}=1e-3, \lambda_{T}=1e-3,\lambda_{Global,Rot}=1e-1, \lambda_{Global,scale}=1.0\)

\begin{figure*}[p]
  \centering

  \includegraphics[height=0.95\textheight]{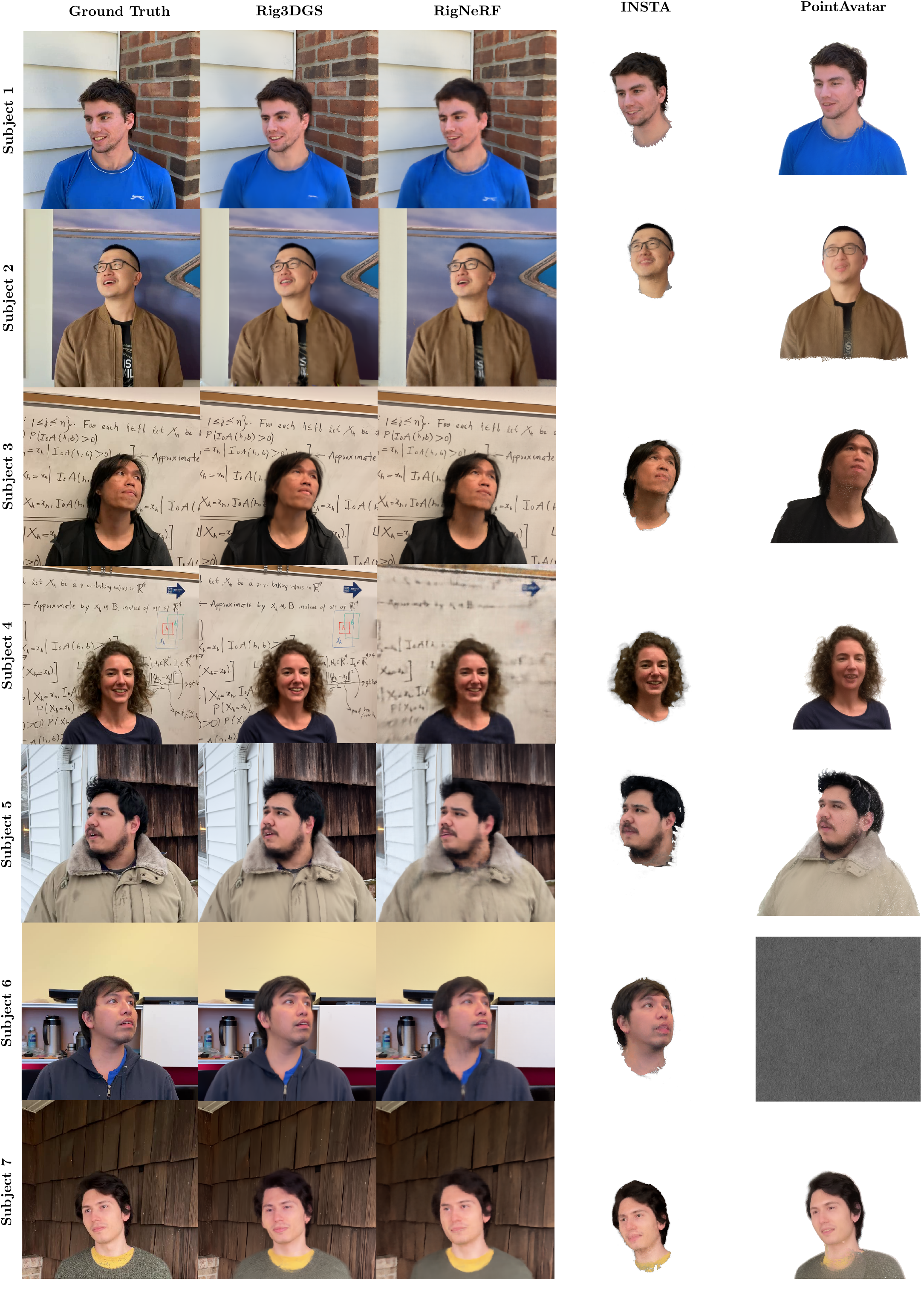}
  \caption{Qualitative comparison of Subjects 1-7 in Setting 1. \MethodName produces full-scene renders with higher-quality facial and background detail than competing baselines. Unfortunately, PointAvatar's results for subject 6 never converged despite 3 different experimental trials.}
  \label{fig:main_res}
\end{figure*}
\section{Results}
\newcommand{\mathcolorbox}[2]{\colorbox{#1}{$\textstyle #2$}}

\begin{table*}[t]
\begin{center}
\small
\scalebox{0.65}{
\begin{tabular}{lcccccccccccccccc}
\toprule
  & \multicolumn{4}{c}{\emph{\MethodName} (Ours)}  &  \multicolumn{4}{c}{\emph{RigNeRF \cite{athar2022rignerf}}}  &  \multicolumn{4}{c}{\emph{INSTA \cite{INSTA:CVPR2023}}} &  \multicolumn{4}{c}{\emph{PointAvatar \cite{Zheng2023pointavatar}}}\\
  \midrule
  Subject (Head Region)   & PSNR $\uparrow$ &SSIM $\uparrow$ & LPIPS $\downarrow$ & DISTS $\downarrow$
   & PSNR $\uparrow$ &SSIM $\uparrow$ & LPIPS $\downarrow$ & DISTS $\downarrow$
  & PSNR $\uparrow$ &SSIM $\uparrow$ & LPIPS $\downarrow$ & DISTS $\downarrow$
  & PSNR $\uparrow$ &SSIM $\uparrow$ & LPIPS $\downarrow$ & DISTS $\downarrow$ \\
\midrule
 Subject 1 &
 \mathcolorbox{pink}{31.70} & \mathcolorbox{pink}{0.9591} & \mathcolorbox{pink}{0.0342} & \mathcolorbox{pink}{0.0845} &
 30.19 & 0.9505 & 0.0469 & 0.1296 &
 29.25 & 0.9530 & 0.0396 & 0.0986 &
 21.77 & 0.9233 & 0.0577 & 0.1292
 \\
Subject 2 &
 \mathcolorbox{pink}{33.10} & \mathcolorbox{pink}{0.9749} & \mathcolorbox{pink}{0.0239} & \mathcolorbox{pink}{0.0739} &
 31.91 &  0.9715  & 0.0266 & 0.0841 &
 30.74 &  0.9694  & 0.0337 & 0.1098  & 
 24.73 &  0.9460  & 0.0435 & 0.1143 
  \\
 Subject 3 &
 26.22 & 0.9447 & 0.0482 & \mathcolorbox{pink}{0.1234} &
  27.65 &  \mathcolorbox{pink}{0.9461}  & \mathcolorbox{pink}{0.0472} & 0.1305 &
\mathcolorbox{pink}{27.77} &  0.9423 & 0.0617 & 0.1629  &
  21.16 &  0.9058  & 0.0751 & 0.1798
\\ 
Subject 4 &
\mathcolorbox{pink}{26.98} & \mathcolorbox{pink}{0.9240} & \mathcolorbox{pink}{0.0577} & \mathcolorbox{pink}{0.1242} &
  26.35 &  0.9100  & 0.0607 & 0.1583 &
  25.34 & 0.9185 & 0.0769 & 0.1718 &
  22.55 & 0.8824  & 0.0792 & 0.1696
 \\ 
Subject 5 &
\mathcolorbox{pink}{30.23} & \mathcolorbox{pink}{0.9464} & \mathcolorbox{pink}{0.0529} & \mathcolorbox{pink}{0.1361} &
  29.69 & 0.9400  & 0.0705 & 0.1771 &
  28.29 &  0.9433  & 0.0615 & 0.1361  &
  22.15 &  0.8902  & 0.0907 & 0.1996 
 \\ 
Subject 6 &
  27.49 & \mathcolorbox{pink}{0.9668} & \mathcolorbox{pink}{0.0352} & \mathcolorbox{pink}{0.0797} &
  27.23 &  0.9605  & 0.0489 & 0.1233 &
\mathcolorbox{pink}{28.58} &  0.9559  & 0.0436 & 0.1011  &
  - & - & - & -
 \\ 
Subject 7 &
\mathcolorbox{pink}{24.07} & 0.9248 & 0.0623 & \mathcolorbox{pink}{0.1250} &
  23.20 &  0.9232  & 0.0688 & 0.1444 &
  23.60 &  \mathcolorbox{pink}{0.9414}  & \mathcolorbox{pink}{0.0619} & 0.1403  &
  22.67 &  0.9180  & 0.0616 & 0.1405
 \\ 
 \bottomrule
\textbf{Average} &
\textbf{28.26} & \textbf{0.9487} & \textbf{0.0450} & \textbf{0.1067} &
  \textbf{28.03} &  \textbf{0.9431}  & \textbf{0.0528} & \textbf{0.1353} &
  \textbf{27.65} & \textbf{0.9463} & \textbf{0.0541} & \textbf{0.1315} &
  \textbf{22.05} &  \textbf{0.9110} & \textbf{0.0680} & \textbf{0.1555}
 \\ 
\bottomrule
\end{tabular}}

\caption{\small{Quantitative results for Subject 1-7's masked head region in Setting 1. Here we calculate PSNR, SSIM, LPIPS, and DISTS using head masks provided by INSTA. Our results are better than RigNeRF \cite{athar2022rignerf}, INSTA \cite{INSTA:CVPR2023} and PointAvatar \cite{Zheng2023pointavatar} most metrics. \mathcolorbox{pink}{\text{Best Score}}}
}

\label{tab:Subjects_metrics_1}
\vspace{-0.3cm}
\end{center}
\end{table*}

\begin{table}[h]
\begin{center}
\small
\scalebox{0.60}{
\begin{tabular}{lcccccccc}
\toprule
  & \multicolumn{4}{c}{\emph{\MethodName} (Ours)} &  \multicolumn{4}{c}{\emph{RigNeRF \cite{athar2022rignerf}}}\\
  \midrule
  Subject (Full Scene)   & PSNR $\uparrow$ &SSIM $\uparrow$ & LPIPS $\downarrow$ & DISTS $\downarrow$
   & PSNR $\uparrow$ &SSIM $\uparrow$ & LPIPS $\downarrow$ & DISTS $\downarrow$ \\
\midrule
 Subject 1 &
 \mathcolorbox{pink}{26.89} & \mathcolorbox{pink}{0.8684} & \mathcolorbox{pink}{0.1583} & \mathcolorbox{pink}{0.1088} &
   25.04 &  0.8083  & 0.2359 & 0.1543
 \\
Subject 2 &
 \mathcolorbox{pink}{25.84} & \mathcolorbox{pink}{0.8665} & \mathcolorbox{pink}{0.1556} & \mathcolorbox{pink}{0.1226} &
    23.31 &  0.7922  & 0.1779 & 0.1297
  \\
 Subject 3 &
 22.44 & \mathcolorbox{pink}{0.8624} & \mathcolorbox{pink}{0.1160} & \mathcolorbox{pink}{0.0923} &
 \mathcolorbox{pink}{22.58} &  0.8230  & 0.1657 & 0.1522
\\ 
Subject 4 &
\mathcolorbox{pink}{23.97} & \mathcolorbox{pink}{0.8670} & \mathcolorbox{pink}{0.1174} & \mathcolorbox{pink}{0.0747} &
  20.34 &  0.6627  & 0.4264 & 0.2955
 \\ 
Subject 5 &
\mathcolorbox{pink}{23.65} & \mathcolorbox{pink}{0.7910} & \mathcolorbox{pink}{0.1723} & \mathcolorbox{pink}{0.0960} &
  22.53 &  0.6770  & 0.3314 & 0.1982
 \\ 
Subject 6 &
\mathcolorbox{pink}{29.94} & \mathcolorbox{pink}{0.9304} & \mathcolorbox{pink}{0.0874} & \mathcolorbox{pink}{0.0674} &
  28.71 &  0.9081  & 0.1266 & 0.1347
 \\ 
Subject 7 &
\mathcolorbox{pink}{22.36} & \mathcolorbox{pink}{0.7503} & \mathcolorbox{pink}{0.1758} & \mathcolorbox{pink}{0.1246} &
  21.22 &  0.6679  & 0.3826 & 0.2379
 \\ 
\bottomrule

\end{tabular}}
\caption{\small{Quantitative results for Subject 1-7's full-scene renders from \MethodName and RigNeRF \cite{athar2022rignerf} in Setting 1.  \mathcolorbox{pink}{\text{Best Score}}}
}
\label{tab:Subjects_metrics_rot_rig}
\vspace{-0.3cm}
\end{center}
\end{table}

\begin{table}[h]
\begin{center}
\small
\scalebox{0.60}{
\begin{tabular}{lcccccccc}
\toprule
  & \multicolumn{4}{c}{\emph{\MethodName} (Ours)} &  \multicolumn{4}{c}{\emph{RigNeRF \cite{athar2022rignerf}}}\\
  \midrule
  Subject (Full Scene)   & PSNR $\uparrow$ &SSIM $\uparrow$ & LPIPS $\downarrow$ & DISTS $\downarrow$
   & PSNR $\uparrow$ &SSIM $\uparrow$ & LPIPS $\downarrow$ & DISTS $\downarrow$ \\
\midrule
 Subject 1 &
 \mathcolorbox{pink}{28.05} & \mathcolorbox{pink}{0.8805} & \mathcolorbox{pink}{0.1394} & \mathcolorbox{pink}{0.0798} &
   26.38 & 0.8203  & 0.2382 & 0.1491
 \\
Subject 2 &
   \mathcolorbox{pink}{25.94} &  \mathcolorbox{pink}{0.8720}  & \mathcolorbox{pink}{0.1729} & \mathcolorbox{pink}{0.1162} &
   22.23 &  0.7900  & 0.2375 & 0.1471
  \\
 Subject 3 &
  \mathcolorbox{pink}{29.83} &  \mathcolorbox{pink}{0.9395} & \mathcolorbox{pink}{0.0721} & \mathcolorbox{pink}{0.0741}  &
  25.11 &  0.8389  & 0.1821 & 0.1641
\\ 
Subject 4 &
  \mathcolorbox{pink}{26.38} &  \mathcolorbox{pink}{0.8814}  & \mathcolorbox{pink}{0.1025} &  \mathcolorbox{pink}{0.0591} &
  22.30 &  0.7096  & 0.4086 & 0.2840 
 \\ 
Subject 5 &
  \mathcolorbox{pink}{25.92} &  \mathcolorbox{pink}{0.8467}  & \mathcolorbox{pink}{0.1419} & \mathcolorbox{pink}{0.0925}  &
  24.12 &  0.7468  & 0.2774 & 0.1790
 \\ 
Subject 6 &
  \mathcolorbox{pink}{28.58} &  \mathcolorbox{pink}{0.9195}  & \mathcolorbox{pink}{0.0961} & \mathcolorbox{pink}{0.0664} &
  25.64 &  0.8269  & 0.2126 & 0.1656
 \\ 
Subject 7 &
  \mathcolorbox{pink}{27.51} &  \mathcolorbox{pink}{0.8169}  & \mathcolorbox{pink}{0.1635} & \mathcolorbox{pink}{0.1171}  &
  25.49 &  0.7339  & 0.3054 & 0.1968
 \\ 
\bottomrule
\end{tabular}}

\caption{\small{Quantitative results for Subject 1-7's full-scene renders from \MethodName and RigNeRF \cite{athar2022rignerf} in Setting 2.  \mathcolorbox{pink}{\text{Best Score}}}
}

\label{tab:Subjects_metrics_pan_rig}
\vspace{-0.3cm}
\end{center}
\end{table}

\begin{table*}[h]
\begin{center}
\small
\scalebox{0.80}{
\begin{tabular}{lcccccccccccccc}
\toprule
  & \multicolumn{4}{c}{\emph{Learanble Prior} (Ours)}  &  \multicolumn{4}{c}{\emph{Fixed Prior}}  &  \multicolumn{4}{c}{\emph{No Prior }}\\
  \midrule
  Subject (Head Region)   & PSNR $\uparrow$ &SSIM $\uparrow$ & LPIPS $\downarrow$ & DISTS $\downarrow$
   & PSNR $\uparrow$ &SSIM $\uparrow$ & LPIPS $\downarrow$ & DISTS $\downarrow$
  & PSNR $\uparrow$ &SSIM $\uparrow$ & LPIPS $\downarrow$ & DISTS $\downarrow$\\
\midrule
 Subject 1 &
 \mathcolorbox{pink}{31.70} & \mathcolorbox{pink}{0.9591} & \mathcolorbox{pink}{0.0342} & \mathcolorbox{pink}{0.0845} &
 30.57 & 0.9500 & 0.04152 & 
 0.0967 & 19.25 & 0.910 & 
 0.0868 & 0.224   
 \\
Subject 5 &
 \mathcolorbox{pink}{30.23} & \mathcolorbox{pink}{0.9464} & \mathcolorbox{pink}{0.0529} & \mathcolorbox{pink}{0.1361} &
 29.81 &  0.9444  & 0.0613  &
 0.1513 &  20.78 & 0.896 & 
 0.09 & 0.199    
  \\
 \bottomrule
\end{tabular}}

\caption{\small{Quantitative ablative study on the efficacy of a learnable deformation prior. We narrow our study to Subjects 1 and 5.  \mathcolorbox{pink}{\text{Best Score}}}}

\label{tab:Subjects_metrics_aba}
\vspace{-0.4cm}
\end{center}
\end{table*}

In this section, we show results of head-pose control, facial expression control, and novel view synthesis using \MethodName. For each scene, the model is trained on a short portrait video captured using a consumer smartphone.

\subsection{Baseline Approaches}
In the context of neural portrait reanimation, the only prior work that offers full control over facial expression, head-pose and viewing direction of the scene is RigNeRF \cite{athar2022rignerf}. RigNeRF represents the dynamic scene as a neural radiance field where the mapping from the deformed space to the canonical space is performed using 3DMM-based deformation. However, since RigNeRF \cite{athar2022rignerf} uses MLPs to model both the deformation and radiance field, it is extremely slow to train and use during inference. At an inference rate of about 8FPS, on a consumer grade GPU, we are about 50 times faster than RigNeRF \cite{athar2022rignerf}. The authors of RigNeRF \cite{athar2022rignerf} kindly provided results of their method on the training data used for this paper. 
A class of methods that are closely related to \MethodName are ones that only control facial expression and head-pose and \textit{do not} model the entire scene. %
Two such prior works are INSTA \cite{INSTA:CVPR2023} and PointAvatar \cite{Zheng2023pointavatar} stand out. INSTA \cite{INSTA:CVPR2023}, creates a controllable human head from monocular smartphone videos. Built on Instant-NGP, a notably fast and high-quality \cite{mueller2022instant} approach to neural rendering, INSTA's \cite{INSTA:CVPR2023} is currently state of the art in terms of quality and speed of human head avatar models. %
Concurrent with INSTA, PointAvatar \cite{Zheng2023pointavatar} represents the scene as a set of spherical points. This model provides valuable insights into the nuanced advantages of 3D Gaussians; consequently, we include it within our analysis. We use the open-source code provided by the authors of INSTA and Pointavatar for our experiments. 

\subsection{Training Data Capture}
We captured the training data using various iPhone models, including a iPhone XR, 12, and 13 Pro Max. The capture process is comprised of two distinct phases. In the first half, subjects are instructed to perform a diverse array of expressions and speech while maintaining a stationary head with the camera to panning around them. Subsequently, in the latter half, the camera is fixed at head-level and subjects are prompted to rotate their heads enacting various facial expressions. All training videos are about 40-70 seconds long, equivalent to approximately 1200-2100 frames, and are down-sampled to a resolution of 512 x 512. We use COLMAP \cite{schoenberger2016sfm} to estimate the Camera parameters and \MethodName's inital point estimate. In order for camera calibration to be accurate, we mask out the dynamic foreground prior to running COLMAP. All models utilize FLAME \cite{FLAME:SiggraphAsia2017} as their 3DMM of choice. For \MethodName we used DECA \cite{DECA} to calculate an initial estimate of FLAME parameters for each video, which we then optimized through standard landmark fitting, using landmarks predicted by 3DDFA-V2 \cite{3DDFA_V2} and camera parameters estimated from COLMAP \cite{schoenberger2016sfm}. While each model utilizes a different subset of FLAME parameters during training, we ensure that expression and shape parameters are shared between models.

\subsection{Evaluation on Test Data}
We assess the performance of \MethodName, RigNeRF \cite{athar2022rignerf}, INSTA \cite{INSTA:CVPR2023}, and PointAvatar \cite{Zheng2023pointavatar} using held-out images extracted from captured video sequences. We test two different settings:
\begin{enumerate}
    \item \textbf{Fixed Camera view, changing expression and head-pose}: This evaluation setting uses a fixed camera view and only varies expressions and head-poses to evaluate their fidelity during reanimation. In this setting we evaluate all methods and the metrics are only calculated on the head-region
    \item \textbf{Fixed Head-pose with changing expression and camera view}: This evaluation setting uses a fixed head-pose and only varies camera views and expressions to evaluate fidelity of view synthesis and facial expressions. In this setting, we only evaluate \MethodName and RigNeRF, since INSTA and PointAvatar do not model the entire scene.
\end{enumerate}
 We use about 30-60 \textit{held-out} images for both settings. In both settings,we measure PSNR, SSIM, LPIPS \cite{perceptual}, and DISTS \cite{DISTS}. While PSNR and SSIM measure pixel-wise accuracy, LPIPS \cite{perceptual} and DISTS \cite{DISTS} measure perceptual accuracy. 
 \subsubsection{Evaluation in Setting 1} In \tab{Subjects_metrics_1} we show quantitative results of the first setting by measuring the metrics only on the head-region with constant view but changing expression and head-pose. As can be seen, across most subjects and metrics, \MethodName outperforms prior work by a large margin. This can also be seen in qualitative results in \fig{main_res}, where \MethodName generates higher quality renders then prior work with greater fidelity to facial expressions and head-pose. 
 While RigNeRF \cite{athar2022rignerf} is able to reasonably reproduce the head-pose and facial expressions, its renders are blurry and lack detail, especially around the eyes and mouth regions. This can be seen most clearly in the case of Subjects 1,3,4,5,6 and 7. In some cases, like that of Subject 4, it fails to reconstruct the background correctly. We attribute this to its use of MLPs that often underfit scenes. Please note, in the original paper \cite{athar2022rignerf}, the authors train RigNeRF \cite{athar2022rignerf} on a resolution of 256x256 while we train on a resolution of 512x512. In \tab{Subjects_metrics_rot_rig}, we quantitatively evaluate the performance of \MethodName and RigNeRF on the full scene. We omit INSTA and PointAvatar since they do not model the full scene. As can be seen, \MethodName outperforms RigNeRF across almost all metrics and on all subjects save the PSNR on Subject 3.  
 Similar to RigNeRF \cite{athar2022rignerf}, while INSTA \cite{INSTA:CVPR2023} is able to generate renders that are reasonably consistent with articulated facial expression and head-pose, it generates artifacts around the mouth, as can be seen in Subjects 2,3,5,6 and 7. artifacts are also present in the hair region (Subjects 2,4,5 and 7) and on accesories such as glasses in the case of Subject 2. Unlike \MethodName, INSTA is unable to model the full scene. 
 While PointAvatar \cite{Zheng2023pointavatar} is able to generate a more complete render of the subject, its results often lack detail and are oversmoothed. It fails to reproduce the mouth region accurately, as can be seen in Subjects 1-5 and 7. The results of PointAvatar also contain dot artifacts due to empty space between points as can seen in Subjects 3,5 and 7. Unfortuantely, PointAvatar did not converge on Subject 6.
 In contrast to prior work, \MethodName is able to model facial expression and head-pose high fidelity to generate high-quality photorealistic renders. 

 \subsubsection{Evaluation in Setting 2} In this setting, we evaluate the ability of \MethodName and RigNeRF \cite{athar2022rignerf} to model facial expressions and the full 3D scene. As can be seen in \tab{Subjects_metrics_pan_rig}, \MethodName outperforms RigNeRF across all metrics and all subjects. This is due to RigNeRF's renders being generally blurry compared to those of \MethodName. In the Supplementary we include more qualitative results to support this claim. 
\subsection{Reanimation with Pose and Expression Control}
\begin{figure*}[h]
  \centering
  \includegraphics[width=1.0\textwidth]{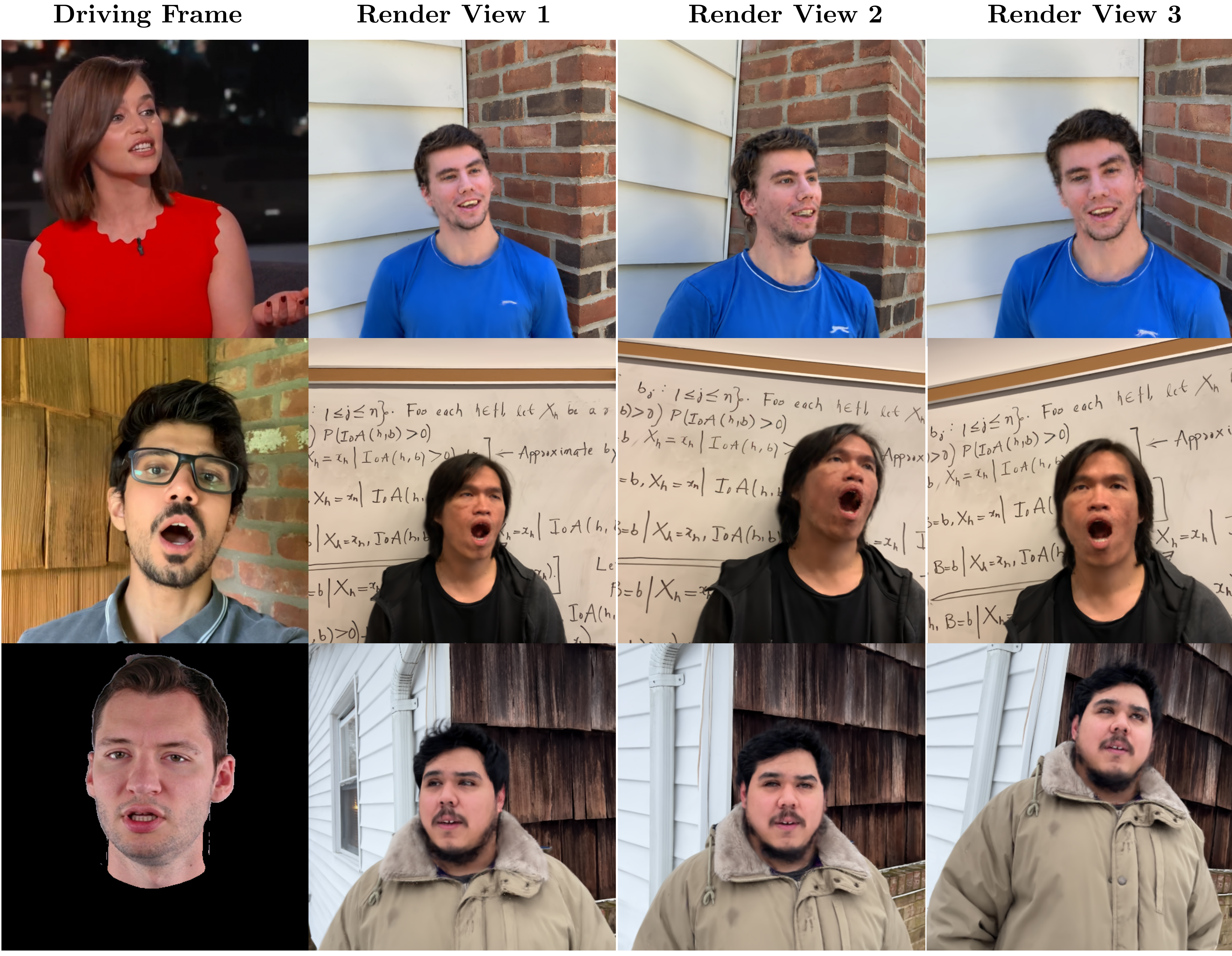}
  \caption{Sample renders of Subjects 1, 3, and 5 reanimated by different expression and head pose donors. We refer the reader to our supplementary video material for a more comprehensive evaluation.}
  \label{fig:nvs_res}
\end{figure*}

In \fig{nvs_res}, we show results of \MethodName being driven by a driving frame in 3 different novel views. As can be seen, across all subjects and driving frames \MethodName is able to reproduce the driving expression and head-pose with high fidelity while simultaneously generating a faithful of the full scene under novel views. We strongly urge the reader to check out the project page for more results on 3DMM driven reanimation and novel view synthesis.

\subsection{Ablating the Deformation Prior} 
\label{sect:aba_def_prior}
In this section, we ablate the learnable deformation prior as defined in \eq{deform_def_eq} and \eq{deform_def_eq_final}. We ablate the following 3 deformation models
\begin{itemize}[leftmargin=*]
    \item No Prior: The deformation is directly predicted by an MLP as follows \(\text{Def}(\xbf_{i,can}, \gammabfiexp,\gammabfipose) = \text{MLP}(\xbf_{i,can}, \gammabfiexp,\gammabfipose)\)
    \item Fixed Prior: The deformation uses a decayed 3DMM-deformation similar to RigNeRF \cite{athar2022rignerf}
    \begin{equation*}
    \begin{split}
        \text{Def}(\xbf_{i,can}, \gammabfiexp,\gammabfipose) &= \text{3DMMDef}(\xbf_{i,can}, \gammabfiexp,\gammabfipose) \\
        &+ \text{MLP}(\xbf_{i,can}, \gammabfiexp,\gammabfipose) 
    \end{split}
    \end{equation*}
    
    \item Learnable Prior Following \eq{deform_def_eq_final}
\end{itemize}

\begin{figure}[h]
  \centering
  \includegraphics[width=1.0\linewidth]{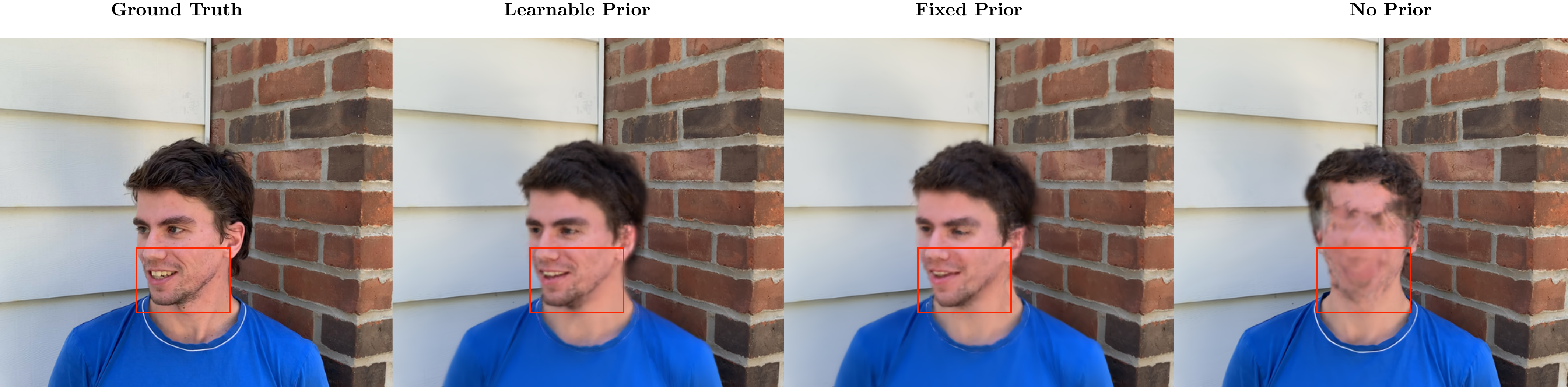}
  \caption{Ablation of the Learnable Deformation Prior. As can be seen, the learnable prior, as defined by \eq{deform_def_eq_final}, is able to model the target expression and head-pose better than the fixed prior (see highlighted regions). The model with no prior fails to reanimate altogether.}
  \label{fig:prior_aba}
\end{figure}

In \fig{prior_aba}, we show qualitative results of these three deformation model for a novel expression and head-pose. As can be seen, the model without a prior is unable to generate any meaningful render. The model with a fixed prior, similar to what RigNeRF \cite{athar2022rignerf} uses, is unable to model fine expressions and gives a somewhat blurry render. In contrast, the proposed learnable deformation prior from \eq{deform_def_eq_final} is able to generate a photorealistic render with high fidelity to the target facial expression and head-pose. In \tab{Subjects_metrics_aba}, we provide a quantitative comparison of between the three deformation model. As can be seen, using a learnable prior, as defined in \eq{deform_def_eq_final} gives us the best results across all metrics on both subjects.

\section{Conclusion}

In this paper we have presented \MethodName, a novel method capable of arbitrary facial expression control and novel view synthesis for portrait videos. \MethodName uses a learnable deformation prior to ensure stability during training and generalization to novel facial expressions, head-pose and viewing direction. \MethodName is also able to model details of the subject's face such as hair and glasses and reproduce them with high fidelity as the video is driven. 
 
However, the problem of controllable human head models with novel view synthesis is far from solved. \MethodName is unable to model strong non-uniform illumination and requires the subject in the portrait video to remain relatively still during capture. We hope to address this in future work. %

\section{Acknowledgements}
This work was supported in part by the CDC/NIOSH through grant U01 OH012476 and a gift from Adobe.

\vspace{3mm}

{
    \newpage
    \bibliographystyle{ieee_fullname}
    \bibliography{refs}
}

\clearpage
\newpage
\section{Supplementary}
\section{Qualitative Comparison on Novel view synthesis}
In this section, provide a qualitative comparison between \MethodName and RigNeRF on novel views and facial expressions as mentioned in Section 4.3.2 of the paper. As can be seen, \MethodName generates significantly sharper render for all subjects across various views and facial expressions.

\begin{figure*}[p]
  \centering

  \includegraphics[height=0.97\textheight]{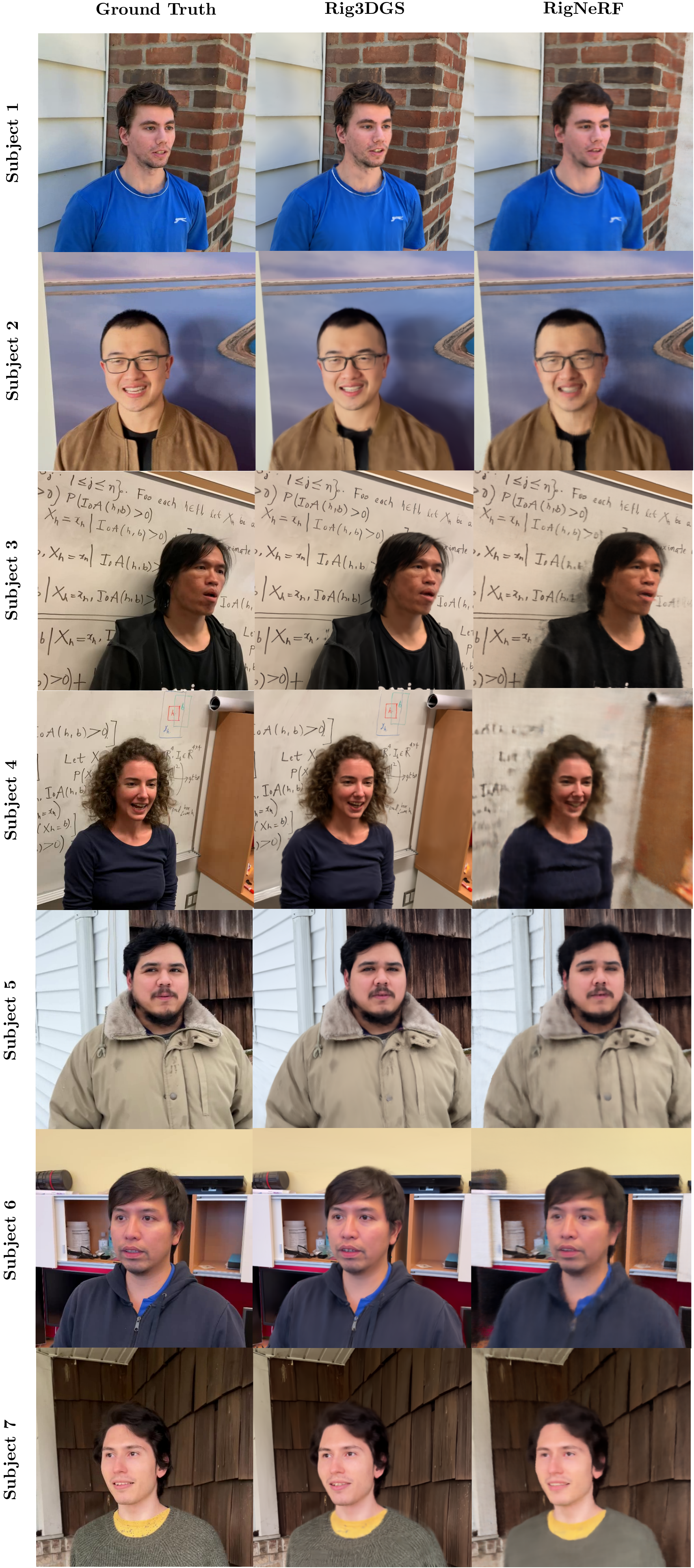}
  \caption{Qualitative comparison of Subjects 1-7 in Setting 2. As can be seen, \MethodName generates higher quality renders than RigNeRF across all subjects.}
  \label{fig:main_res_setting_2}
\end{figure*}

\section{Experimental Details}
All models of \MethodName were trained for 4 hours on an A5000 GPU with a batch-size of 1 for 40k-60k iterations. Densification and pruning of points was done until 10k iterations. The learning rate for the gaussian positions is exponentially decayed from 7e-4 to 8e-6 over 20k iterations. The learning rate for the deformation triplane, \(F\) and associated MLP's (\(\text{G}, T, R^{*}, S^{*} \)) are decayed from 5e-4 to 1e-5 over 40k iterations. \(R^{*} \text{ and } S^{*}\) have 4 layers while \(\text{G} \text{ and } T\) use 2 layers, all MLP's use a   \texttt{Softplus(\(\beta\)=10.0)} activation function. In Eqs 5,8 and 9 \(\xbf_{i,can}\) is positionally encoded with 10 frequencies while \(\text{DWAvg}(\{\delta \vbf_{j}(\gammabfiexp,\gammabfipose\}\})\) is encoded with 4 frequencies. 

\end{document}